\documentclass[letterpaper]{article} 
\usepackage[]{aaai24}  
\usepackage{times}  
\usepackage{helvet}  
\usepackage{courier}  
\usepackage[hyphens]{url}  
\usepackage{graphicx} 
\usepackage{natbib}  
\usepackage{caption} 
\usepackage{algorithm}
\usepackage{algorithmic}

\usepackage{newfloat}
\usepackage{listings}
\DeclareCaptionStyle{ruled}{labelfont=normalfont,labelsep=colon,strut=off} 
\floatstyle{ruled}
\newfloat{listing}{tb}{lst}{}
\floatname{listing}{Listing}
\usepackage{graphicx}
\usepackage{amsmath}
\usepackage{amssymb}
\usepackage{multirow}

\newcommand{\ve}[1]{{\mathbf #1}} 

\newcommand{\spa}[1]{{\mathbb #1}}

\usepackage{comment}
\usepackage{amssymb}
\usepackage{pifont}

\usepackage[utf8]{inputenc} 
\usepackage[T1]{fontenc}    
\usepackage{url}            
\usepackage{booktabs}       
\usepackage{amsfonts}       
\usepackage{nicefrac}       
\usepackage{microtype}      
\usepackage{xcolor}         

\usepackage{multicol}
\usepackage{multirow}
\usepackage{caption}
\usepackage{algorithm}
\usepackage{algorithmic}

\usepackage{color,colortbl}
\definecolor{Gray1}{gray}{0.9}
\definecolor{Gray2}{gray}{0.99}

\DeclareMathOperator*{\argmin}{arg\,min}

\title{Selective Feature Adapter for Dense Vision Transformers}
\author{Xueqing Deng$^{1}$, Qi Fan$^{2}$\thanks{Work done during internship at ByteDance}, Xiaojie Jin$^{2}$, Linjie Yang$^{1}$, Peng Wang$^{1}$ \\
$^{1}$ ByteDance, USA, $^{2}$ Hong Kong University of Science and Technology  \\ 
}

\begin{document}
\maketitle

\begin{abstract}
    Fine-tuning pre-trained transformer models, e.g., Swin Transformer~\cite{liu2022swin}, are successful in numerous  downstream 
    for dense prediction vision tasks. However, one major issue is the cost/storage of their huge amount of parameters, which becomes increasingly challenging to handle with the growing amount of vision tasks. 
    In this paper, we propose an effective approach to alleviate the issue, namely selective feature adapter (SFA). It achieves state-of-the-art (SoTA) performance under any given budget of trainable parameters, and demonstrates comparable or better performance than fully fine-tuned models across various dense tasks. 
    Specifically, SFA consists of external adapters and internal adapters which are sequentially operated over a transformer model. For external adapters, we properly select the places and amount of additional multilayer perception (MLP).  For internal adapters, we transform a few task-important parameters inside the transformer, which are automatically discovered through a simple yet effective lottery ticket algorithm. Our experiments show that the dual adapter module, a.k.a SFA, is essential to achieve the best trade-off on dense vision tasks, such as segmentation, detection and depth-estimation, outperforming other adapters with a single module.
\end{abstract}

\section{Introduction}
\label{sec:intro}


With the ever-growing capacity of model and size of data, pre-trained vision transformers (ViT)~\cite{vit2021iclr,liu2022swin} on large datasets such as ImageNet21K~\cite{ridnik2021imagenet}, JFT-300M~\cite{sun2017revisiting} or JFT-3B~\cite{zhai2022scaling} have shown great success and achieve SoTA performance in various downstream vision tasks with smaller data size through a full model finetuning. 
For example, representative works with this paradigm are presented in segmentation with ADE20K~\cite{zhou2017ade20k}, detection with COCO~\cite{lin2014coco} and depth/normal estimation~\cite{ranftl2021vision} with NYUv2~\cite{silber2012nyuv2} etc. 
However, large models cost huge online storage in memory of GPUs when it is deployed in practice which is expensive to serve. For example, the SoTA SwinTransformer Huge has 3 billion parameters~\cite{liu2022swin} requiring 6GB using float16 in memory during the serving. When using a separately finetuned model ($X$), the total serving size increase linearly with the number of vision tasks, i.e. $nX$, can become prohibitively expensive when serving a large number of models.

\begin{figure}[!t]
\centering
\includegraphics[width=1.0\linewidth]{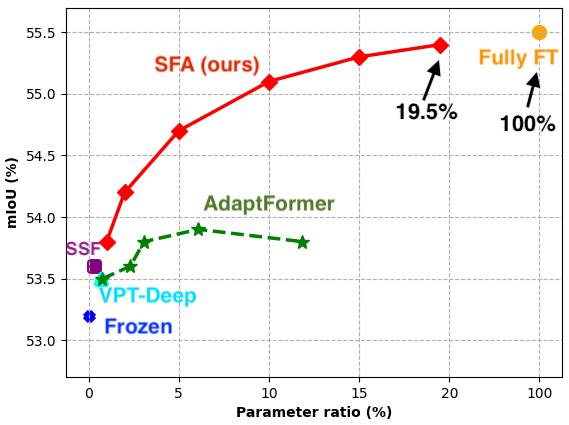}
\vspace{-20pt}
\caption{Performance on ADE20K with the backbone of SwinV2-Base pretrained on ImageNet22K. With $\leq$ 1\% parameters, SFA achieves the best adaptation performance. With 19.6\% parameters, it is comparable with fully fine-tuning (Full-FT).  }
\vspace{-10pt}
\label{fig:teaser}
\end{figure}

To alleviate such an issue, several works have been proposed for classification such as using a totally frozen backbone~\cite{lin2022could}, manually selecting some internal bias for finetuning~\cite{jia2022vpt}, or adding some external module~\cite{chen2022adaptformer}. 
In this paper, we are particularly interested in large model performance on dense prediction vision tasks such as segmentation, detection, etc. 
After investigating all the existing works on a segmentation task, we find that there are several remaining problems in practice: 1) these strategies sacrifice too much performance for reducing the parameter cost, which is important for accuracy-sensitive applications. 2) it is not easy to adjust the budget of trainable parameters, where the availability of multiple options is important when the deployment hardware is changed. Therefore, in this paper, we propose a selective feature adapter (SFA) which solve both issues simultaneously with a single strategy. 

As shown in Fig.~\ref{fig:teaser}, for segmentation, SFA achieves the best trade-off between performance and varying budget of trainable parameters, outperforming previous SoTA strategies such as VPT~\cite{jia2022vpt} and Adaptformer~\cite{chen2022adaptformer}.  
Furthermore, we are able to match the performance of the fully fine-tuned model with only $\sim$ 20\% of its parameters. 
Inspired from AdaptFormer~\cite{chen2022adaptformer}, we first introduce an external feature adapter, while propose a novel design that boosts its performance on dense vision tasks. However, with the increase of external selected parameters, we observe its performance is saturated with a gap from the Full-FT model since these adapters need to be learned from scratch. Therefore, following the principle of "Not All Parameters Are Equal"\cite{frankle2018lottery}, we hypothesize that some features inside are essential for adapting a pre-trained transformer, and further propose an internal selected adapter, which successfully closes the performance gap.

\begin{figure*}[!t]
\centering
\includegraphics[width=0.83\linewidth]{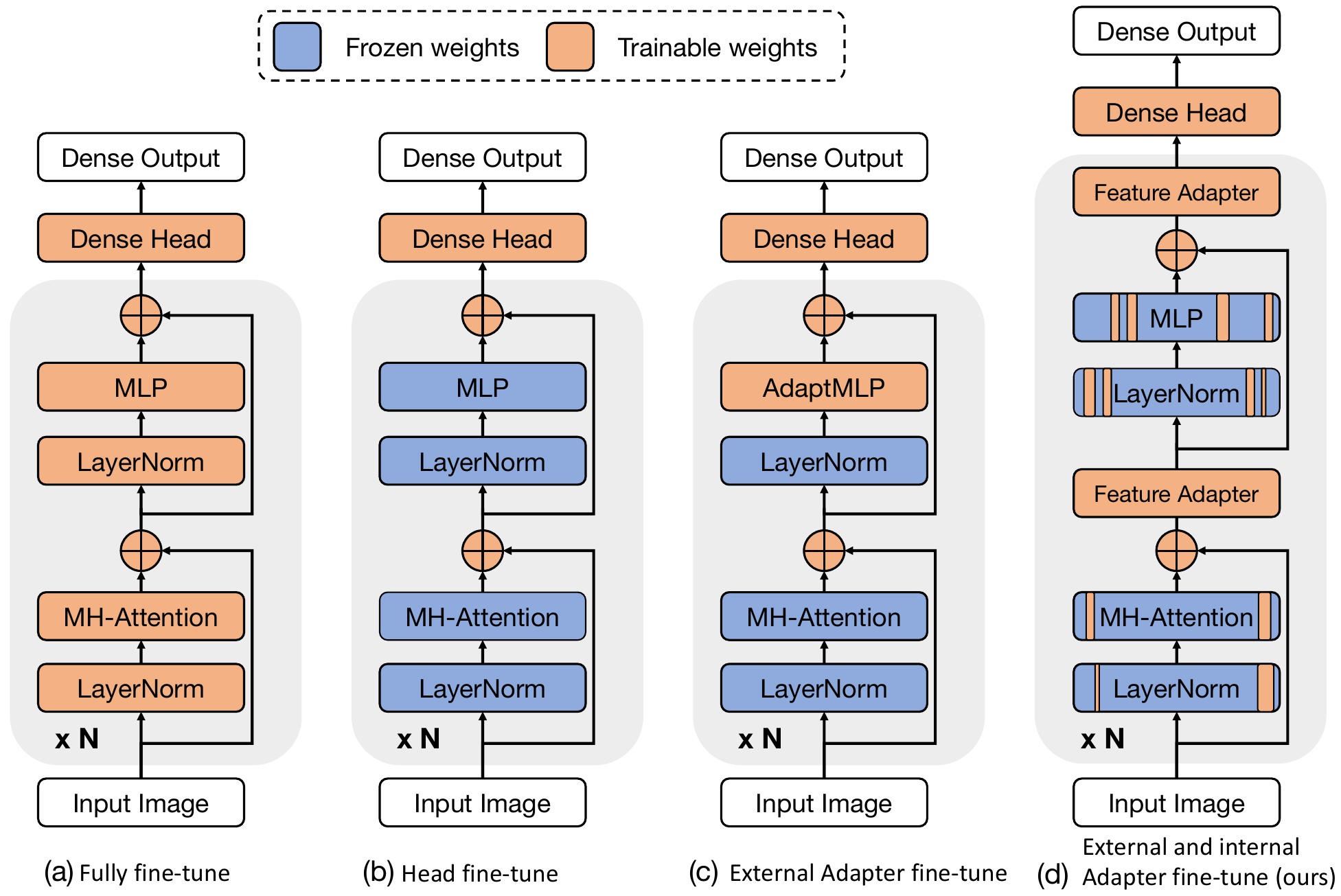}
\vspace{-2pt}
\caption{Fine-tuning methods for adaptation. (a) All model parameters are trained; (b) Task specific head are trained; (c) External only adapters; (d) SFA (Ours) with both internal and external adapters. }
\vspace{-10pt}
\label{fig:method_comparison}
\end{figure*}

Fig.~\ref{fig:method_comparison} illustrates that, unlike previous approaches that either fine-tune only the head of a network or add an external adapter, SFA contains two adapters - an internal adapter that transforms essential selected features and external adapters stacked above both multi-head attention and MLP. 
In our experiments, we validate the superiority of SFA design on multiple transformers and tasks, including segmentation, detection, instance segmentation, and depth estimation. We also show that SFA selected adapters can generalize across difference datasets, such as from ADE20k~\cite{zhou2017ade20k} to cityscape~\cite{cordts2016cityscapes}, demonstrating the effectiveness of our approach.

In summary, our contributions are in three aspects:
\begin{enumerate}
    \item We propose selective feature adapter (SFA), a dual feature adapter that effectively adapts a giant vision transformer to dense prediction vision task. It achieves SoTA performance across various budgets of trainable parameters.
    \item We formulate such a problem of adapting a giant vision transformer to a dense prediction vision task as a adapter search problem (Sec.~\ref{sec:method}), which helps discover SFA.
    \item We conducted extensive ablation experiments to validate each design of SFA and demonstrate its superiority over multiple dense prediction tasks.
\end{enumerate}
Our implementation will be released along with the publication of this paper. 


\section{Related Works}
\label{sec:related}





\paragraph{Large pre-trained transformers.} Transformers have demonstrated outstanding results on natural language processing~\cite{devlin2018bert, lan2020albert, liu2019roberta} and computer vision tasks~\cite{vit2021iclr,xie2021segformer,wang2021crossformer,lee2021vitgan,touvron2021deit}. State-of-the-art models often increase their size to leverage various architectural designs for better performance. 
In the meantime, larger-scale datasets~\cite{ridnik2021imagenet, sun2017revisiting} are also proposed to pre-train these models which are able to transfer to the downstream tasks with a significant performance boost. For example, in natural language processing, BERT~\cite{devlin2018bert} achieves great success with parameter size of 340M. While GPT-3~\cite{brown2020gpt3} is even much larger which grows to 175 billion parameters and is trained with nearly a trillion words.
Similar trends are happening in computer vision, ViT~\cite{vit2021iclr} proposes the first vision transformer architecture with few modifications by splitting the input images into patches. 
Later, various designs considering the pyramid of vision architectures are proposed with great success, e.g. Swin Transformers~\cite{liu2022swin, liu2021swinv1} with ImageNet22K, MAE ~\cite{he2022mae} and BEiT~\cite{bao2022beit} with large unlabelled datasets. 
In summary, the dramatically increased model sizes for both language and vision tasks raise challenges of fine-tuning on the whole set of model parameters~\cite{ding2022delta} with growing downstream tasks which can cause expensive storage costs.

\paragraph{Multi-task/Incremental learning.} One straightforward way is adding extra heads for any additional downstream task by freezing the backbone parameters~\cite{kendall2018multi,lin2022could}. 
However, these methods often lead to a degradation of the performance on the target task, compared with a well-finetuned model. 
Other more complex strategies could modify architectures and learn partial parameters~\cite{guo2020spottune} to balance the performance across different tasks, such as the research conducted in the fields of multi-task learning~\cite{vandenhende2021multi} or incremental learning~\cite{de2021continual}. Nevertheless, these approaches have to consider the history of included tasks, therefore adding significant extra cost to model training. 

\paragraph{Prompt/Parameter efficient tuning.} Another trend in transformer-based works is to optimize parameter efficiency by incorporating small trainable modules, referred to as "prompts", that use a minimal number of parameters (e.g., 0.1\% as seen in \cite{zhai2019large}) to fine-tune large pre-trained models for downstream tasks.
Similarly, these works, as surveyed in~\cite{qiao2022survey}, are first proposed in the language domain, which can be categorized into token-based and network-based methods. Token-based methods~\cite{zhang2022dart,tam2021adapet}  propose to prepend several learnable prefix vectors/tokens to the projected tokens from the inputs to  their attention. 
While network-based methods often leverage more complex modules inside of the transformers, e.g. reprojected attention layer~\cite{stickland2019pal}, low-rank approximation (LoRA)~\cite{hu2022lora},
bottleneck adapter modules~\cite{houlsby2019parameter}, and even a unified framework for multi-downstream tasks~\cite{he2022towards}. 


Recently, researchers also extend these ideas to the vision domain for visual classification. 
For example, VPT~\cite{jia2022vpt} propose to adopt a small amount of extra parameters(1\%) in input space for finetuning specific tasks with a frozen transformer. 
SSF~\cite{lian2022ssf} further extends the prompting idea by learning a scaling and a shift operation for the outputs of each layer. AdaptFormer~\cite{chen2022adaptformer} proposes a trainable MLP module which is inplace in parallel beside each MLP layer inside of the transformer, yielding significant improvement in video action recognition. NOAH~\cite{zhang2022noah} propose to use a neural architecture search (NAS) algorithm to find the optimal prompt adapter modules with a search space consisting of the modules from existing works~\cite{chen2022vitadapter, hu2022lora, jia2022vpt}.
Though performing well on classification or on a relatively small multiple task benchmark with  VTAB-1k~\cite{zhai2019large}, when we try to transfer these algorithms on other popular dense prediction tasks, such as ADE20k and COCO, we find there is still a significant performance gap towards the fully-finetuned models. 
ViT-Adapter~\cite{chen2022vitadapter} proposed a convolution plus MLP adapter architecture, which can be jointly tuned with transformers, yielding strong performance on dense prediction tasks. However, it is not designed for parameter efficient training, which is still sub-optimal when we perform it with a frozen backbone. For dense prediction in SFA, we utilize an optimization target and discover a simpler but more optimized adapter. 

\section{Method}
\label{sec:method}

In this section, we discuss our proposed method in detail. Specifically,
we introduce the framework of the Selective Feature Adapter (SFA) and its formulation in Section \ref{subsec:sfa}. We then delve into the specifics of our dual adapter architecture and outline how to find an optimal adapter that balances efficiency and effectiveness. 





\subsection{Selective Feature Adapter}
\label{subsec:sfa}
As presented in the introduction (Sec.~\ref{sec:intro}), we are interested in finding a good feature adapter based on a given transformer and a given vision task. Here, we first formulate such a problem as an objective function, and then introduce the ideas of selective feature adapter (SFA) to optimize it.

Formally, given a pre-trained transformer model $\mathcal{T}$, and a down-stream dataset $\{\ve{X}_i, \ve{Y}_i\}_{i}^{N}$, where $\ve{X}_i$ and  $\ve{Y}_i$ represents the input and label of a data instance. 
Our target is to find an optimal feature adapter $\mathcal{F}$ respect to $\mathcal{T}$ under any given budget of trainable parameters $\beta$, which can be formulated as,
\begin{equation}
\begin{aligned}
    \max_{\mathcal{F}} \quad & \mathcal{A}_{val}\left(\theta^*(\mathcal{F}(\mathcal{T}))\right) \\ 
    \textrm{s.t.} \quad & \theta^*(\mathcal{F}(\mathcal{T})) = \argmin_{\theta} \quad L_{train}(\theta(\mathcal{F}(\mathcal{T})), \\
    & \|\theta(\mathcal{F}(\mathcal{T}))\| \leq \beta \|\theta(\mathcal{T})\|
\end{aligned}
\label{eqn:objective}
\end{equation}
where $\mathcal{A}_{val}$ is the accuracy on the validation set.  $\theta^*(\mathcal{X})$  is the optimal parameters of the network $\mathcal{X}$ on the training set using the given loss $L_{train}()$. $\theta(\mathcal{X})$ represents the network parameters, and  $\|\theta(\mathcal{X})\|$ counts the amount of parameters. Here, we require that the trainable parameter of the adapter $\mathcal{F}$ should not exceed a fraction, denoted as $\beta$, of the total parameters in the pre-trained transformer. 
With such a definition, our ultimate goal is to find an adapter $\mathcal{F}$ such that the accuracy on validation is maximized, and simultaneously the amount of trainable parameters of the adapter $\|\mathcal{F}\|$ is minimized. Here, for simplicity, we drop $\ve{X}$ and $\ve{Y}$ in the formula for loss and accuracy calculation.

Next, we decompose the adapter $\mathcal{F}$ into an external adapter $\mathcal{F}_e$ and an internal adapter $\mathcal{F}_i$, where the external adapter is added as intermediate layers into the transformer, and the internal adapter finds the most important parameters inside the transformer that need to be tuned. In our design, $\mathcal{F}_e$ and $\mathcal{F}_i$ are sequentially used to adapt $\mathcal{T}$, which can be formulated as,
\begin{equation}
\begin{aligned}
    \mathcal{F}(\mathcal{T}) &= \mathcal{F}_i(\mathcal{F}_e(\mathcal{T})), \\
    \theta(\mathcal{F}) &= \{\theta(\mathcal{F}_e), \theta(\mathcal{F}_i)\}, \\
    \|\theta(\mathcal{F}_e)\| &\leq \beta_e \|\theta(\mathcal{T})\|, \|\theta(\mathcal{F}_i)\| \leq \beta_i \|\theta(\mathcal{T})\|
\end{aligned}
\label{eqn:separate}
\end{equation}
where $\beta_e = \rho\beta$ and $\beta_i = (1-\rho)\beta$. $\rho$ is a hyperparameter to distribute the total budget for the two components, and is set to $0.5$ in our experiments. 
In the following, we will elaborate our motivation and details of constructing each adapter with respect to our objective. 




\begin{figure}[!t]
\centering
\includegraphics[width=0.9\linewidth]{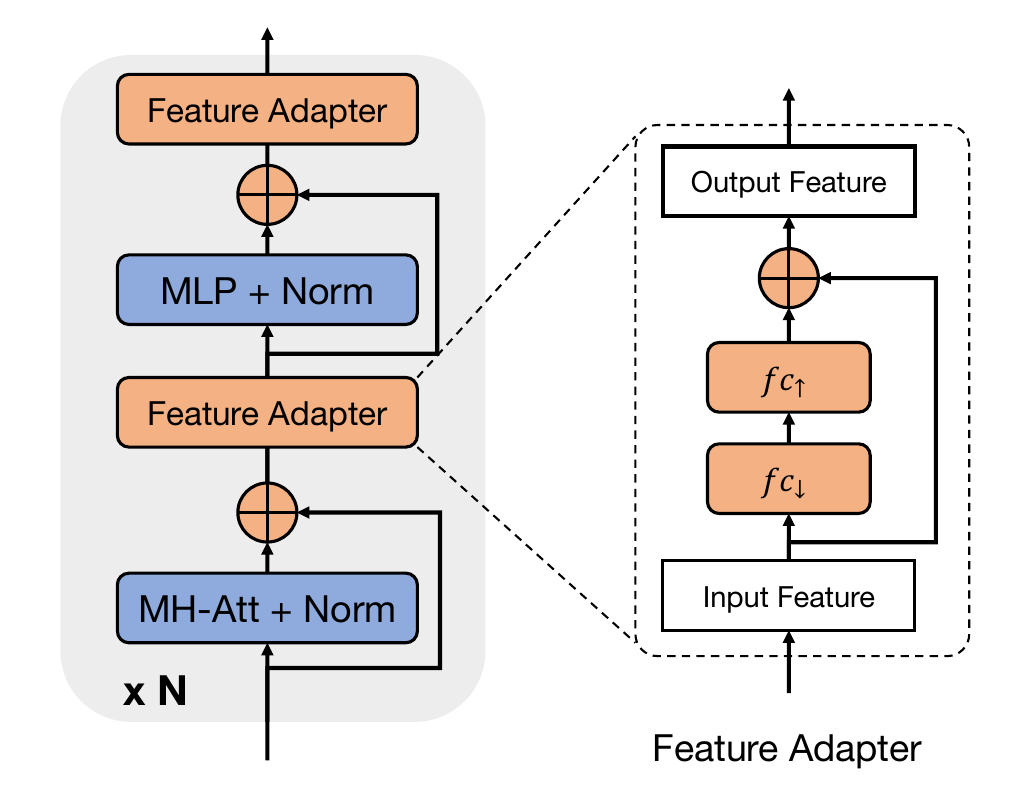}
\caption{External Feature adapter $\mathcal{F}_e$, where the adapters are inserted right after the multi-head attention (MH-Att) and multilayer perception (MLP). Norm represents the layer norm operation.}
\vspace{-10pt}
\label{fig:external}
\end{figure}

\subsection{External Feature Adapter $\mathcal{F}_e$}
\label{subsec:external}
As discussed in Sec.~\ref{sec:related}, external adapters do not modify the parameters inside of the transformer $\mathcal{T}$, which is a commonly adopted strategy due to its simplicity~\cite{chen2022adaptformer}. 
Here in SFA, we adopt this idea and re-design the adapter formula to benefit our dense prediction tasks as shown in Fig.~\ref{fig:external}. 

Specifically, suppose $\ve{x}_l \in \spa{R}^{D_l}$ is the feature after the multi-head attention (MH-Att) and multilayer perception (MLP) in a transformer block, our external feature adapter is defined as,

\begin{equation}
    \begin{aligned}
        \ve{x}_l' = \ve{x}_l + \text{FC}_{d\rightarrow D_l}(\text{ReLU}(\text{FC}_{D_l\rightarrow d}(\ve{x}_l)))
    \end{aligned}
\label{eqn:block}
\end{equation}

From the formula, we borrow the architecture of AdaptMLP as designed in AdaptFormer~\cite{chen2022vitadapter} due to its effectiveness and efficiency. 
However, our adapter contains two key differences that are important for the final performance under various vision tasks. 
First, rather than adding the external adapter as a side path along with the internal MLP with a scaling factor~\cite{chen2022adaptformer}, we consider a sequential plus residual strategy without the scaling (Eqn.~\ref{eqn:block}). Such an architecture on one hand increases the depth of the network which could be potentially more effective as stated in previous arts~\cite{eldan2016power}. On the other hand, the residual connections, as introduced in ResNet~\cite{he2016deep}, help maintain the original transformer features from the frozen backbone, making the network easier to learn and converge. 
The second difference is that we place adapters after both the MH-Att and the MLP, which further increases the network depth and boosts the performance. Although we added multiple components in the transformer layers, the additional computation latency is marginal (310ms for a SwinV2-Base transformer vs 380ms for SwinV2-Base with adapters on a \textbf{Nvidia V100} for a 10\% adapter). 

In our experiments (Sec.~\ref{sec:exp}), we demonstrate our external adapter $\mathcal{F}_e$ works more effectively on dense vision tasks, e.g. segmentation, than AdaptMLP.
One may think about adding more stacks of $\mathcal{F}_e$ at each location for better performance. However, we have not observed significant benefits with this setting for our tasks which shows some performance drop due to overfitting.

\begin{figure*}[!t]
\centering
\includegraphics[width=0.95\linewidth]{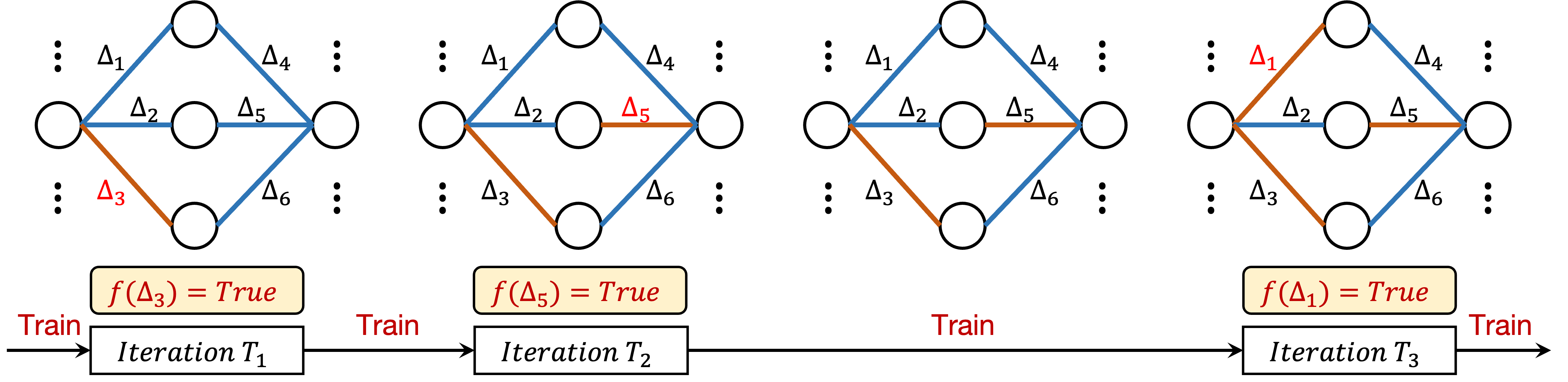}
\vspace{-5pt}
\caption{Process of discovering the trainable parameters for internal adapter $\mathcal{F}_i$. Illustrated with a two-layer MLP, where the trainable parameters are gradually discovered and trained through the training process. It is important to note that the trainable parameters are discovered and selected only at specific iterations, i.e., $T_1$, $T_2$ and $T_3$ in this example. The backbone weights are always trained once they are selected.}
\vspace{-5pt}
\label{fig:internal}
\end{figure*}

\subsection{Internal Feature Adapter $\mathcal{F}_i$}
\label{subsec:internal}
In this section, we introduce $\mathcal{F}_i$ which further adapts a small subset of parameters from the pretrained transformer $\mathcal{T}$. 
This is motivated by our observation in practice, as also introduced in Sec.~\ref{sec:intro}, that on dense prediction tasks like segmentation, the performance with simple external adaptors is saturated with a gap from the fully-finetuned model, even with a very large MLP dimension $d$. In order to match the fully-finetuned performance, we may need to adjust some important feature representations inside the pretrained model. 

Specifically, let $\mathcal{F}_i = \{\ve{f}_{ml}, \ve{f}_{al}\}_{l=1}^{N}$ be the selected feature channels in the transformer backbone. $\ve{f}_{al}$ and $\ve{f}_{ml}$ are the selected features in the MH-Att  and MLP  at layer $l$, respectively.
To find an optimal $\mathcal{F}_i$, our method shares a similar strategy as lottery ticket hypothesis~\cite{frankle2018lottery} and progressive network pruning~\cite{lin2020dynamic,zhu2017prune}, where we search for a small group of tunable parameters in a large pretrained model, keeping the majority of parameters fixed. 

In Fig.~\ref{fig:internal}, we illustrate such a process with a single two-layer MLP $\ve{m}$ for simplicity. The process with MH-Att is following the same pipeline. 
Suppose its original pre-trained weight is $\theta_0(\ve{m})$, and the selected feature parameters are $\ve{f}$.
The winning ticket with iterative optimization process~\cite{zhu2017prune,lin2020dynamic} contains $n$ rounds of following steps: 
\begin{itemize}
    \item Train the $\ve{m}$ for $t = T/n$ iterations, arriving at parameters $\theta_t(\ve{m})$.
    \item Calculate the gradient magnitude of each parameter yielding a feature difference vector 
    \begin{align}
        \Delta_{t} = |\theta_0(\ve{m}) - \theta_t(\ve{m})|. 
        \label{eqn:gradient}
    \end{align}
    \item Select top $\frac{\beta_i}{n}$ percentile while excluding ones already chosen in $\ve{f}$. Formally, for a particular parameter $p \in \ve{m}$, it is selected when,
    \begin{align}
        f(\Delta_p) = R(\Delta_p | \Delta_t) < \frac{\beta_i}{n}\|\theta_t(\ve{m})\|,
        \label{eqn:internal}
    \end{align}
    where $R(x | \ve{x})$ is a decreasing ranking function returning rank of $x$ in $\ve{x}$.
    \item Append the selected parameters to $\ve{f}$, and reset the other parameters to $\theta_0$. 
\end{itemize}
In practice, our feature selection process goes along with the finetuning using the same training recipe as retraining. After the training is done, we may directly use the trained model for evaluation without re-training which is usually required in neural-architecture-search (NAS)~\cite{zoph2016neural}. Here, $n$ is determined by training step size $n = T/s$, and $s$ is a hyperparameter, which is optimized based on a validation set. $s$ is set to 2000 in our experiments.

\subsection{Understand SFA w.r.t. the Objective}

After such a process, we find the sparse feature adapter $\mathcal{F}_i$ is complementary to the external one $\mathcal{F}_e$,  which further improves the performance of the adapted model on dense prediction tasks under a given budget.
Finally,  we summarize the overall SFA algorithms in the pipeline in Alg.~\ref{alg:algorithm}.

\begin{algorithm}[tb]
\caption{SFA}
\label{alg:algorithm}
\textbf{Input}: 
Train/Val dataset $D$; 
Pre-trained transformer with parameter $\theta(\mathcal{T})$. 
Budget of parameter $\beta$. 
Number of iterations $T$ and step size $s$. \\
\textbf{Output}: Trained Adapted Transformer $\mathcal{F}(\mathcal{T})$.\\
\begin{algorithmic}[1] 
\STATE Split $D$ into $D_{train}$ and $D_{val}$.
\STATE Split $\beta$ in $\{\beta_e, \beta_i\}$ by a portion $\rho$.
\STATE Choose $d$ in $\mathcal{F}_e$, \text{s.t} $\|\theta(\mathcal{F}_e)\|\approx\beta_e \|\theta(\mathcal{T})\|$
\STATE Let $t=1$ and $n=T/s$.
\WHILE{$t < T$}
    \STATE Compute gradient $\delta_t(\mathcal{F}(\mathcal{T}))$ and $\delta_t(\mathcal{T})$ with $L_{train}$ in Eqn.~\ref{eqn:objective}.
    \STATE Update adapter $\theta_t(\mathcal{F}(\mathcal{T}))$ for both internal and external adapters.
    \STATE Accumulate gradient $\delta_t(\mathcal{T})$ to $\Delta_t(\mathcal{T})$.
        \IF{$t \% s = 0$}
            \STATE Find partial internal adapter $\mathcal{F}_i^t$ with portion of $\frac{\beta_i}{n}$ in Eqn.~\ref{eqn:internal}.
            \STATE Append the $\mathcal{F}_i^t$ to $\mathcal{F}_i$.
            \STATE Reset $\Delta_t(\mathcal{T}) = 0$
        \ENDIF
\ENDWHILE
\end{algorithmic}
\end{algorithm}

Recalling the objective in Eqn.~\ref{eqn:objective},
for external adapter $\mathcal{F}_e$, we considered the architectures from ViT-Adaptor~\cite{chen2022vitadapter}, Adaptformer~\cite{chen2022adaptformer}, and ours in Fig.~\ref{fig:external}. We select current architectures based on validation accuracy $\mathcal{A}_{val}()$ on segmentation, and our experiments demonstrate that it generalizes well across various dense tasks. 

Regarding the internal adapter $\mathcal{F}_i$, as in Eqn.~\ref{eqn:gradient}, the trainable features are selected using first-order gradients generated solely with $L_{train}$, without considering validation accuracy $\mathcal{A}_{val}$. 
However, our experiments show that this approach performs well on different tasks due to its underlining regularization strategy, as discussed in~\cite{frankle2018lottery}. 

While it is possible to adopt second-order optimization methods, such as those used in DARTs~\cite{liu2018darts}, to improve the selection, many one-shot NAS works~\cite{yu2019autoslim,xie2018snas} have shown only marginal improvement with significantly increased training costs. Additionally, for learning the adapters, there may be more effective schedulers~\cite{yin2020sooner} or advanced lottery strategies~\cite{hoefler2021sparsity} than the proposed iterative selection process. Although we believe that these strategies can be integrated into our adapter search, we keep the simple strategy to showcase the potential of our approach.




Lastly, it is possible to reverse the order of adapters for improved performance, i.e., using $\mathcal{F} = \mathcal{F}_e(\mathcal{F}_i(\mathcal{T}))$. However, this approach requires training a model without $\mathcal{F}_e$ to search for $\mathcal{F}_i$ first, and then fine-tuning again with $\mathcal{F}_e$ while freezing $\mathcal{F}_i$, which doubles the training cost. Despite this, our experiments indicate that this approach achieves similar performance to the original SFA design. Thus, we retain the original SFA design for this work.

\begin{table*}[] 
  \caption{Comparison on ADE20K with multi-scale mIoU. We considered backbones of SwinV2-Base pre-trained with imagenet1K (S-B-1K) and imagenet22K (S-B-22K), SwinV2-Large pre-trained with imagenet22K (S-B-22K). Param \% indicates the portion of learnable parameters.} 
  \vspace{-5pt}
    \centering
    \resizebox{0.68\linewidth}{!}{%
      \begin{tabular}{l|ccc|cc}
        \toprule
        Method & Param \% & S-B-1K & S-B-22K & Param \% & S-L-22K \\
        \midrule
        Frozen~\cite{lin2022could} & 0  & 51.4 & 53.2 & 0 & 54.4 \\
        SSF~\cite{lian2022ssf}& 0.3 & 51.7 & 53.6 & 0.2 & 54.8 \\
        VPT-Deep~\cite{jia2022vpt} & 0.7 & 51.6 & 53.5 & 0.5 & 54.9 \\
        LoRA ~\cite{hu2022lora} & - & - & - & 1.2 & 54.8 \\
        AdaptFormer~\cite{chen2022adaptformer} & 13.5 & 51.8 & 53.8 & 9.2 & 55.2 \\
        
         SFA-S (ours) & 4.8 & \textbf{52.2} & \textbf{54.7} & 4.5 & \textbf{55.9}  \\
       
        \midrule
        Fully fine-tune  & 100 & 52.9 & 55.5 & 100 & 56.5 \\
         SFA-B (ours) & 19.5 & 52.8 & 55.4 & 19.0 & 56.5 \\
        \bottomrule
      \end{tabular}}
\vspace{-10pt}
\label{table:ade20k}
\end{table*}

\section{Experiments}
\label{sec:exp}

In this section, we verify SFA on dense prediction tasks including semantic segmentation, object detection and instance segmentation, and depth estimation.

\subsection{Datasets and Implementation Details}
\label{sec:datasets}
We select SwinV2 series~\cite{liu2022swin} as our backbone to perform experiments across all tasks. 
The variants include SwinV2-Base with 88M parameters and SwinV2-Larg with 197M parameters. Here, we are not able to perform experiments using even larger models such SwinV2-G since there is no official released version.
More details are shown below.

\vspace{+3pt}
\noindent{\bf Semantic Segmentation.} 
We adopt semantic segmentation dataset ADE20K~\cite{zhou2017ade20k} for both the ablation study and comparison study. ADE20K consists of 150 semantic categories with 20K training images and 2K validation images. Mask2former~\cite{cheng2021mask2former} with a six-block pixel detector is used as the segmentation head. For the backbone, we investigate these variants including SwinV2-Base pretrained on ImageNet1K (S-B-1K) and ImageNet22K (S-B-22K), and SwinV2-Large on ImageNet22K (S-L-22K). We develop our experiments on MMSegmentation~\cite{mmseg2020} using same experimental setting as in~\cite{lin2022could}. We evaluate the effectiveness by reporting the mIoU score with multi-scale test augmentation on the validation set. 

\vspace{+3pt}
\noindent{\bf Object Detection and Instance Segmentation.}
In this part, we consider MS COCO~\cite{lin2014coco} dataset. It contains 80 categories, with 118K training images and 5K validation images. Mask R-CNN~\cite{he2017maskrcnn} with the neck of FPN~\cite{lin2017fpn}/BiFPN~\cite{tan2020bifpn} and Cascade Head~\cite{cai2018cascade} is selected for implementation. S-B-22K is adopted as the backbone. We report the average precision (AP) of the box (detection) and mask (segmentation) on the validation set. We conduct our experiments on MMDetection~\cite{mmdetection} following settings in~\cite{lin2022could}.

\vspace{+3pt}
\noindent{\bf Depth Estimation.}
For depth estimation, we consider two benchmark datasets, NYUv2~\cite{silber2012nyuv2} and KITTI~\cite{geiger2012kitti}. NYUv2 covers 464 indoor scenes, with 24K training images and 654 testing images and KITTI includes diverse outdoor self-driving scenes, with 23K training images and 697 testing images. We set the maximum depth range as 10m and 80m for NYUv2 and KITTI respectively. The depth estimation head~\cite{xie2022revealing} consists of three deconvolutional layers (with BN and ReLU) and an output convolution layer. Then we select two backbones including S-B-22K and S-L-22K. We implement the experiments using MM-Depth-Estimation~\cite{xie2023mmdepth} and evaluate with rooted mean square error (RMSE).

\begin{table*}[!t]
  \caption{Comparison study on MS COCO object detection and instance segmentation reported with box and mask (shown in parenthesis) AP. The backbone is S-B-22K. FPN~\cite{lin2017fpn}, BiFPN~\cite{tan2020bifpn}, Cascade~\cite{cai2018cascade}}
  \centering
  \resizebox{0.68\linewidth}{!}{%
      \begin{tabular}{l|c|ccc}
        \toprule
        Method  & Param \% &FPN & BiFPN & BiFPN w. Cascade \\
        \midrule
        Frozen~\cite{lin2022could} &0  & 45.0 (41.1) & 51.9 (46.0)  & 53.8 (46.7) \\
        SSF~\cite{lian2022ssf}  &0.3& 46.5 (42.5) & 51.9 (45.7) & 53.9 (46.7) \\
        VPT-Deep~\cite{jia2022vpt}&0.7 & 46.3 (42.2) & 51.9 (45.6) & 54.0 (46.7) \\
        AdaptFormer~\cite{chen2022adaptformer}&13.5  & 47.0 (42.8) & 52.0 (45.7) & 54.0 (46.6) \\
        SFA-S (ours)  &4.8  & \bf{50.2 (44.0)}& \bf{52.1 (45.7)}  & \bf{54.0 (46.8)}  \\
        \midrule
        Fully fine-tune &100 & 51.9 (45.7) & 52.3 (45.7) & 54.3 (46.9) \\
        SFA-B (ours) &19.5 & 51.4 (45.1) & 52.2 (45.9) & 54.2 (46.9) \\
        \bottomrule
      \end{tabular}
  }
\label{table:coco}
\end{table*}

\subsection{Comparison Study}
In this section, we present our experimental results on the selected downstream tasks mentioned in Sec.~\ref{sec:datasets}. We compare SFA to other SoTA adapters such as SSF~\cite{lian2022ssf}, VPT-Deep~\cite{jia2022vpt}, and AdaptFormer~\cite{chen2022adaptformer}. 
Two variants of SFA were evaluated in this study, namely SFA-small (SFA-S) and SFA-base (SFA-B), with parameter ratios around 5\% and 20\%, respectively. 
It is worth noting that SFA-S is used to compare against other adapters, while SFA-B was designed to explore the necessary trainable parameter ratio required to match the performance of the fully fine-tuned model.

\begin{table*}[!t] 
  \caption{Comparison study on NYUv2 and KITTI for depth estimation reported with  RMSE. S-B-22K is defined the same as Tab.~\ref{table:ade20k}. } 
    \centering
      \vspace{-5pt}
         \resizebox{0.68\linewidth}{!}{%
      \begin{tabular}{l|ccc|ccc}
        \toprule
        &  \multicolumn{3}{c|}{S-B-22K} & \multicolumn{3}{c}{S-L-22K} \\
        Method  & Param \% & NYUv2 & KITTI & Param \%& NYUv2 & KITTI \\
        \midrule
        Frozen~\cite{lin2022could} &0 & 0.387 & 2.631 &0  & 0.383 & 2.574 \\
        SSF~\cite{lian2022ssf}&0.3  & 0.369 & 2.527 &0.2& 0.362 & 2.420 \\
        VPT-Deep~\cite{jia2022vpt} &0.7  & 0.371 & 2.552 &0.5& 0.368 & 2.484 \\
        AdaptFormer~\cite{chen2022adaptformer} &13.5 & 0.363 & 2.480 &9.2& 0.359 & 2.411 \\
        SFA-S (ours) & 4.8& \bf{0.347}  & \bf{2.384} &4.5 & \bf{0.345} & \bf{2.302} \\
        
        \midrule
        Fully fine-tune &100  & 0.335 & 2.240 & 100&0.334 & 2.150 \\
        SFA-B (ours)& 19.5&0.340 & 2.302 &19.0& 0.339 & 2.215 \\
        \bottomrule
      \end{tabular}}
\label{table:depth}
\end{table*}

\begin{table}[!t] 
  \caption{Comparison of the external adapter between $\mathcal{F}_e$ (ours) and AdaptFormer under different middle dimensions $d$ w.r.t. mIoU scores on ADE20K.} 
    \centering
    \resizebox{1.05\linewidth}{!}{%
      \begin{tabular}{c|c|cccc}
        \toprule
        & Dim & mIoU & Speed & \# Param (\%) & Memory \\
        \midrule
        Frozen~\cite{lin2022could} & - & 53.2 & 0.31s & 0 & 4.42G \\
        \midrule
        \multirow{5}{*}{AdaptFormer} & 32 &  53.5 & 0.33s & 0.77M (0.9) & 8.52G \\
         & 64 &  53.6 & 0.34s & 1.54M (1.8) & 8.67G \\
         & 128 & 53.8 & 0.36s & 3.08M (3.5) & 8.82G \\
         & 256 & 53.9 & 0.38s & 6.09M (6.9) & 8.97G \\
         & 512 & 53.9 & 0.40s & 11.86M (13.4) & 9.12G \\
        \midrule
        \multirow{4}{*}{$\mathcal{F}_e$ (ours)} & 32 & 53.5 & 0.34s & 1.54M (1.8) & 8.67G \\
        & 64 & 53.8 & 0.36s & 3.08M (3.5) & 8.82G \\
        & 128 & 54.2 & 0.38s & 6.09M (6.9) & 8.97G \\
        & 256 & 54.2 & 0.40s & 11.86M (13.4) & 9.12G \\
        \bottomrule
      \end{tabular}}
\vspace{-10pt}
\label{table:external}
\end{table}

\noindent{\bf {Semantic Segmentation: }} Tab.~\ref{table:ade20k} shows results on ADE20K, where
SFA-S outperforms other adapters across all backbones with a reasonable small parameter ratio. 
Specifically, in first 4 lines, we present the results form Frozen backbone~\cite{zhai2022scaling}, SSF~\cite{lian2022ssf}, VPT~\cite{jia2022vpt} and LoRA~\cite{hu2022lora}. Though they have extremely small ratio of parameters, the gap towards fully finetuned model is significant, i.e. 54.8 (SSF) $\rightarrow$ 56.5 for SWin-L-22K. As for Adaptformer~\cite{chen2022adaptformer} (our major baseline), it closes the gap a bit with mIoU of 55.2, but with a significant amount of additional parameters (13.5\%). SFA, on the other hand, reached significant better results (55.9) with much smaller parameter ratio (4.8\%). 
Furthermore, SFA-B is able to match mIoU scores as fully finetuning with around 19.0\% trainable parameters on SWin-L. 

We conduct the experiments to verify the effectiveness of our selected weights can achieve comparable results with less than 20\% parameters achieving the performance of fully fine-tuning across different datasets as shown in Tab.~\ref{tab:citys}. In details, with the selected weights from the backbone of S-B-1K on ADE20K denoted as SFA-B*, we then perform finetuning on Cityscapes. It is noted that we turn on finetuning on the selected weights instead of directly tranfering the trained weights. Besides, we provide the results for the transfer ability of the selected weights across different segmentation tasks, please refer to our supplementary for more details.


\vspace{+3pt}
\noindent{\bf{Object Detection and Instance Segmentation}} Tab.~\ref{table:coco} shows results on MS COCO, which draws similar observations as segmentation other than SFA combined with different task heads. 
As shown in the table, SFA shows a significant advantage when  FPN~\cite{lin2017fpn} is adopted. 
In detail, the improvement margin is particularly large reaching a gap of 3.2 of box AP and 1.2 of mask AP compared to AdaptFormer. 
The gaps narrow down when the head complexity is significantly increased, for example, when using BiFPN~\cite{tan2020bifpn} or combining BiFPN and Cascade~\cite{cai2018cascade} together. 
This is because when heads becomes more complex, it starts replacing the functionality in the backbone. Therefore, we concludes that SFA is more effective with stronger backbone.  
Similarly, we find SFA can achieve comparable results of fully fine-tuning with around 20\% tunable parameters. We omit the experiments of Swin-L due to the limitation of the computational resource.

       
\vspace{+3pt}

\noindent{\bf{Depth Estimation}} Tab.~\ref{table:depth} shows the results for depth estimation. We evaluate SFA for depth estimation on two benchmark datasets with S-B-22K and S-L-22K respectively. 
As shown in the table, SFA is able to achieve the best performance on both datasets with similar conclusion. SFA-S with S-B-22K using trainable weights can reduce RMSE from 0.363 to 3.347 on NYUv2, and from 2.480 to 2.384 on KITTI respectively compared with AdaptFormer~\cite{chen2022adaptformer}. While increasing the backbone size with S-L-22K, SFA-S can still remain comparable performance gain. 
However, the SFA-B, though provides much better performance than SFA-S, still has some gap towards fully finetuned model, meaning that the regression task is more difficult for adapters, and we would like to study this in our future work.

In conclusion, our experiments demonstrate the effectiveness of SFA. On one hand, reaching good reasults need more parameters than that of SSF and VPT-Deep, while increasing trainable parameter size only does not guarantee performance gain using AdaptFormer. 
On the other hand, we try to answer ``how many parameters are need for an adapter to match the performance of fully fine-tuning''. 
Based on our results, we found it could be around 20\% for tasks of detection/segmentation, while probably more for depth estimation.






\subsection{Ablation Study}
In this section, we aim to validate the performance of individual components of the proposed SFA. To this end, we report mIoU scores on semantic segmentation tasks using S-B-22K as the backbone on the ADE20K dataset following the settings in Sec.~\ref{sec:datasets}. 
Here, we put only the important ablations, in our supplementary, we conduct more ablation including selection criteria and manners of adapter selection, yielding the best one presented in this paper.

\begin{figure*}[!t]
\centering
\includegraphics[width=1\linewidth]{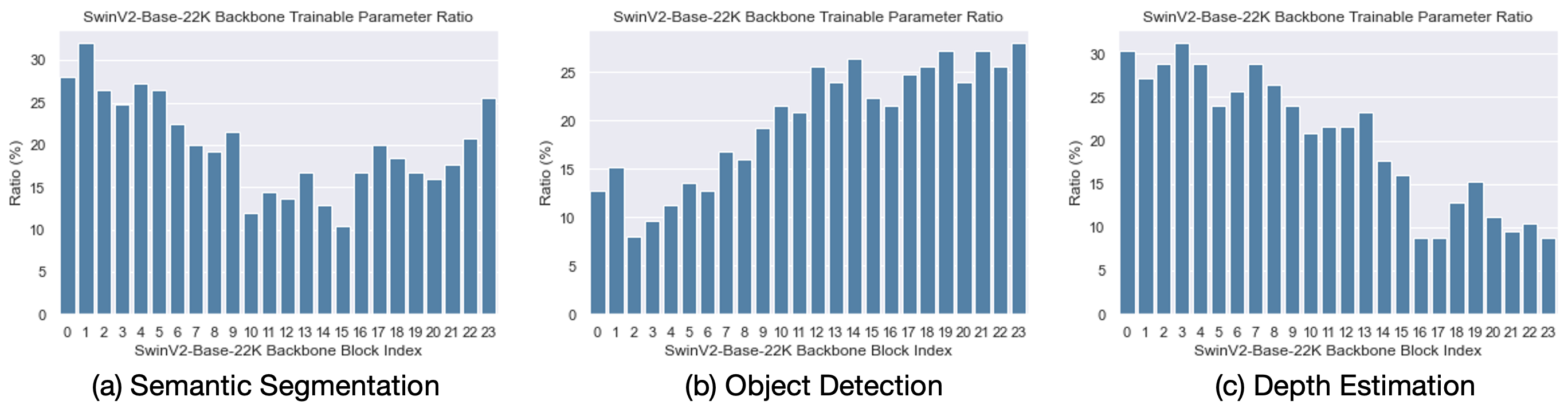}
\vspace{-20pt}
\caption{Selected weights distribution of SFA-B on three dense prediction tasks. The horizontal axis means the layer id, and the vertical axis means the ratio of the selected parameters inside of all. }
\vspace{-10pt}
\label{fig:selected_weights}
\end{figure*}

\vspace{+3pt}
\noindent{\bf{Finding optimal external feature adapter $\mathcal{F}_e$}.} In Tab.~\ref{table:external}, we compare our proposed external adapter $\mathcal{F}_e$(Sec.~\ref{subsec:external}) in terms of memory cost, inference speed and accuracy under various dimensional choices, i.e. $d$ in Eqn.~\ref{eqn:block}, to the external adapter proposed in AdaptFormer. 
 We observe that under ours consistently outperform Adaptformer. 
In addition, increasing the middle dimension can lead to higher mIoU scores while saturated until a dimension, i.e. 256 for Adapformer and 128 for ours. 
At last, we select 64 for SFA-S and 128 for SFA-B as middle-dimension numbers to perform our experiments.

\vspace{+3pt}
\noindent{\bf{Training steps $s$ for $\mathcal{F}_i$ }.} We explore optimization training steps $s$ in $\mathcal{F}_i$ in a wide range as shown in Tab.~\ref{table:taining_steps}. Experimental results show that $s$=2000 yield the best performance and hence it is adopted across all experiments.
The larger and smaller training steps result in  worse performance and the zero training step leads to the worst performance due to the gradient is too random for feature chosen in this case.

\begin{table}[!t] 
  \caption{Ablation study of training steps $s$ in $\mathcal{F}_i$. mIoU scores on ADE20K are reported.} 
    \centering
    \vspace{-5pt}
    \resizebox{1.04\linewidth}{!}{%
      \begin{tabular}{c|cccccccc}
        \toprule

        Step &0& 1000 & 2000 & 3000 & 5000 & 10000 & 20000 & 40000  \\
        \midrule
        mIoU & 53.8&54.6 & \textbf{54.7} & 54.6 & 54.5 & 54.3 & 54.2 & 54.0  \\
        \bottomrule
      \end{tabular}}
\vspace{-5pt}
\label{table:taining_steps}
\end{table}

\vspace{+3pt}
\noindent{\bf{Necessary of dual adapter.}} We conduct experiments to find out the roles of internal and external adapters by varying trainable parameter budgets. 
The results are shown in Tab.~\ref{table:combination}, we can see the combination of $\mathcal{F}_e$  and $\mathcal{F}_i$ can yield better performance than $\mathcal{F}_i$ only under any budget. 
Furthermore, when compared to the external only setting, dual adapter with 5\% adapter ratio reaches 54.7 outperforming 54.2 with 6.9\% adapter ratio using $\mathcal{F}_e$ with $d=128$ in Tab.~\ref{table:external}.

\begin{table}[!t] 
  \caption{Ablation study on adapters under different trainable parameters budgets. mIoU on ADE20K are reported.} 
    \centering
    \resizebox{0.9\linewidth}{!}{%
      \begin{tabular}{l|cccccc}
        \toprule
        \# Param  & 0.86M & 2.58M & 4.3M & 8.6M & 12.9M & 17.2M \\
        Ratio  & 1\% &2\% &5\% &10\%& 15\%&20\% \\
        
       \midrule
       $\mathcal{F}_i$  & 53.5 & 53.8 & 54.3 & 54.7 & 54.8 & 55.0 \\
       $\mathcal{F}_i$ + $\mathcal{F}_e$ & \textbf{53.7} & \textbf{54.2} & \textbf{54.7} & \textbf{55.1}  & \textbf{55.3} & \textbf{55.4} \\
        \bottomrule
      \end{tabular}}
\vspace{-8pt}
\label{table:combination}
\end{table}

\vspace{+3pt}
\noindent{\bf{Distribution of selected internal adapters}.}
We apply our Internal Adapter $\mathcal{F}_i$ (SFA-B model) with the S-B-22K backbone on three dense prediction tasks.
Fig.~\ref{fig:selected_weights} shows the distribution of the selected weights for each task, along with the corresponding layer-wise weight ratio. For object detection, SFA selects more high-level parameters for training, as detailed object boundary feature is not that critical for box annotation.
While, for semantic segmentation, SFA tends to select more balanced features in both low-level and high-level, since semantic segmentation requires both high-level semantic information about the objects and their context in the image, as well as low-level information about the location and appearance of individual pixels. For depth estimation, more low-level parameters are selected due to the fact that accurate depth map requires more local structural details. In conclusion, these results support our intuition that adapters should be task-dependent and the effectiveness of SFA. 

\vspace{+3pt}
\noindent{\bf{Generalization across backbone and datasets.}} 
For backbone, we also consider the ViT backbone~\cite{vit2021iclr} in Vit-Adapter~\cite{chen2022vitadapter} for classification. 
However, we find ViT does not perform as well as SwinV2 for dense tasks. Therefore, due to space limit, we leave its results in the supplementary where SFA also performs the best in terms of our trade-off criteria.

For dataset, we consider transfer the learned SFA using SwinV2 from ADE20k to Cityscapes~\cite{cordts2016cityscapes}, and in~Tab.\ref{tab:citys}, we show that without re-selection, our adapters also works well.

\vspace{+3pt}
\noindent{\bf{Serving storage cost}.} Suppose for $n$ datasets, we need $nX$ storage with Fully-FT. Given SFA, for a task $i$, we only store the difference on selected weights, which is a small portion $p_i$ of the original, yielding a total storage of $\sum_ip_iX \ll nX$.

In addition, based on our study of dataset transfer, SFA enables continuous  adapter selection, where previously selected adapters can be integrated to the whole backbone for later adaptation. This will lead to a further reduction of the cost.

\begin{table}[]
\centering
\caption{Comparison study on Cityscapes. SFA-B* denotes the result using  selected adapters on ADE20K. }
\vspace{-5pt}
\resizebox{0.6\linewidth}{!}{%
\begin{tabular}{l|cc}
\toprule
Method          & Params \% & mIoU    \\
\midrule
Frozen          & 0         & 77.9     \\
SFA-B*          & 19.5      & 82.9   \\
Fully fine-tune & 100       & 83.3    \\
\bottomrule
\end{tabular}}
\vspace{-10pt}
\label{tab:citys}
\end{table}



\section{Conclusion}
This paper proposes SFA, an effective feature adapter for adapting any giant transformer to dense vision tasks. SFA consists of an internal adapter that automatically selects important domain-specific feature parameters inside the transformer, and an external adapter that further improves performance. Our analysis of the selected features reveals task-dependent adaptation patterns. We hope this strategy motivates researchers to investigate the features inside giant models with respect to vision tasks. In future work, we plan to study cross-task feature sharing to further reduce the need of trainable parameters for new vision tasks.

{\small
\bibliography{egbib}

\begin{thebibliography}{61}
\providecommand{\natexlab}[1]{#1}

\bibitem[{Bao et~al.(2022)Bao, Dong, Piao, and Wei}]{bao2022beit}
Bao, H.; Dong, L.; Piao, S.; and Wei, F. 2022.
\newblock {BE}iT: {BERT} Pre-Training of Image Transformers.
\newblock In \emph{ICLR}.

\bibitem[{Brown et~al.(2020)Brown, Mann, Ryder, Subbiah, Kaplan, Dhariwal,
  Neelakantan, Shyam, Sastry, Askell et~al.}]{brown2020gpt3}
Brown, T.; Mann, B.; Ryder, N.; Subbiah, M.; Kaplan, J.~D.; Dhariwal, P.;
  Neelakantan, A.; Shyam, P.; Sastry, G.; Askell, A.; et~al. 2020.
\newblock Language models are few-shot learners.
\newblock \emph{NeurIPS}.

\bibitem[{Cai and Vasconcelos(2018)}]{cai2018cascade}
Cai, Z.; and Vasconcelos, N. 2018.
\newblock Cascade R-CNN: Delving Into High Quality Object Detection.
\newblock In \emph{CVPR}.

\bibitem[{Chen et~al.(2019)Chen, Wang, Pang, Cao, Xiong, Li, Sun, Feng, Liu,
  Xu, Zhang, Cheng, Zhu, Cheng, Zhao, Li, Lu, Zhu, Wu, Dai, Wang, Shi, Ouyang,
  Loy, and Lin}]{mmdetection}
Chen, K.; Wang, J.; Pang, J.; Cao, Y.; Xiong, Y.; Li, X.; Sun, S.; Feng, W.;
  Liu, Z.; Xu, J.; Zhang, Z.; Cheng, D.; Zhu, C.; Cheng, T.; Zhao, Q.; Li, B.;
  Lu, X.; Zhu, R.; Wu, Y.; Dai, J.; Wang, J.; Shi, J.; Ouyang, W.; Loy, C.~C.;
  and Lin, D. 2019.
\newblock {MMDetection}: Open MMLab Detection Toolbox and Benchmark.
\newblock \emph{arXiv preprint arXiv:1906.07155}.

\bibitem[{Chen et~al.(2022)Chen, Chongjian, Tong, Wang, Song, Wang, and
  Luo}]{chen2022adaptformer}
Chen, S.; Chongjian, G.; Tong, Z.; Wang, J.; Song, Y.; Wang, J.; and Luo, P.
  2022.
\newblock AdaptFormer: Adapting Vision Transformers for Scalable Visual
  Recognition.
\newblock In \emph{NeurIPS}.

\bibitem[{Chen et~al.(2023)Chen, Duan, Wang, He, Lu, Dai, and
  Qiao}]{chen2022vitadapter}
Chen, Z.; Duan, Y.; Wang, W.; He, J.; Lu, T.; Dai, J.; and Qiao, Y. 2023.
\newblock Vision Transformer Adapter for Dense Predictions.
\newblock In \emph{ICLR}.

\bibitem[{Cheng et~al.(2022)Cheng, Misra, Schwing, Kirillov, and
  Girdhar}]{cheng2021mask2former}
Cheng, B.; Misra, I.; Schwing, A.~G.; Kirillov, A.; and Girdhar, R. 2022.
\newblock Masked-attention Mask Transformer for Universal Image Segmentation.
\newblock In \emph{CVPR}.

\bibitem[{Contributors(2020)}]{mmseg2020}
Contributors, M. 2020.
\newblock {MMSegmentation}: OpenMMLab Semantic Segmentation Toolbox and
  Benchmark.
\newblock \url{https://github.com/open-mmlab/mmsegmentation}.

\bibitem[{Cordts et~al.(2016)Cordts, Omran, Ramos, Rehfeld, Enzweiler,
  Benenson, Franke, Roth, and Schiele}]{cordts2016cityscapes}
Cordts, M.; Omran, M.; Ramos, S.; Rehfeld, T.; Enzweiler, M.; Benenson, R.;
  Franke, U.; Roth, S.; and Schiele, B. 2016.
\newblock The cityscapes dataset for semantic urban scene understanding.
\newblock In \emph{Proceedings of the IEEE conference on computer vision and
  pattern recognition}, 3213--3223.

\bibitem[{De~Lange et~al.(2021)De~Lange, Aljundi, Masana, Parisot, Jia,
  Leonardis, Slabaugh, and Tuytelaars}]{de2021continual}
De~Lange, M.; Aljundi, R.; Masana, M.; Parisot, S.; Jia, X.; Leonardis, A.;
  Slabaugh, G.; and Tuytelaars, T. 2021.
\newblock A continual learning survey: Defying forgetting in classification
  tasks.
\newblock \emph{IEEE transactions on pattern analysis and machine
  intelligence}.

\bibitem[{Devlin et~al.(2018)Devlin, Chang, Lee, and
  Toutanova}]{devlin2018bert}
Devlin, J.; Chang, M.-W.; Lee, K.; and Toutanova, K. 2018.
\newblock Bert: Pre-training of deep bidirectional transformers for language
  understanding.
\newblock In \emph{ACL}.

\bibitem[{Ding et~al.(2022)Ding, Qin, Yang, Wei, Yang, Su, Hu, Chen, Chan, Chen
  et~al.}]{ding2022delta}
Ding, N.; Qin, Y.; Yang, G.; Wei, F.; Yang, Z.; Su, Y.; Hu, S.; Chen, Y.; Chan,
  C.-M.; Chen, W.; et~al. 2022.
\newblock Delta tuning: A comprehensive study of parameter efficient methods
  for pre-trained language models.
\newblock \emph{arXiv preprint arXiv:2203.06904}.

\bibitem[{Dosovitskiy et~al.(2021)Dosovitskiy, Beyer, Kolesnikov, Weissenborn,
  Zhai, Unterthiner, Dehghani, Minderer, Heigold, Gelly, Uszkoreit, and
  Houlsby}]{vit2021iclr}
Dosovitskiy, A.; Beyer, L.; Kolesnikov, A.; Weissenborn, D.; Zhai, X.;
  Unterthiner, T.; Dehghani, M.; Minderer, M.; Heigold, G.; Gelly, S.;
  Uszkoreit, J.; and Houlsby, N. 2021.
\newblock An Image is Worth 16x16 Words: Transformers for Image Recognition at
  Scale.
\newblock In \emph{ICLR}.

\bibitem[{Eldan and Shamir(2016)}]{eldan2016power}
Eldan, R.; and Shamir, O. 2016.
\newblock The power of depth for feedforward neural networks.
\newblock In \emph{Conference on learning theory}.

\bibitem[{Frankle and Carbin(2019)}]{frankle2018lottery}
Frankle, J.; and Carbin, M. 2019.
\newblock The lottery ticket hypothesis: Finding sparse, trainable neural
  networks.
\newblock In \emph{ICLR}.

\bibitem[{Geiger, Lenz, and Urtasun(2012)}]{geiger2012kitti}
Geiger, A.; Lenz, P.; and Urtasun, R. 2012.
\newblock Are we ready for Autonomous Driving? The KITTI Vision Benchmark
  Suite.
\newblock In \emph{CVPR}.

\bibitem[{Guo et~al.(2019)Guo, Shi, Kumar, Grauman, Rosing, and
  Feris}]{guo2020spottune}
Guo, Y.; Shi, H.; Kumar, A.; Grauman, K.; Rosing, T.; and Feris, R. 2019.
\newblock SpotTune: Transfer Learning Through Adaptive Fine-Tuning.
\newblock In \emph{CVPR}.

\bibitem[{He et~al.(2022{\natexlab{a}})He, Zhou, Ma, Berg-Kirkpatrick, and
  Neubig}]{he2022towards}
He, J.; Zhou, C.; Ma, X.; Berg-Kirkpatrick, T.; and Neubig, G.
  2022{\natexlab{a}}.
\newblock Towards a Unified View of Parameter-Efficient Transfer Learning.
\newblock In \emph{ICLR}.

\bibitem[{He et~al.(2022{\natexlab{b}})He, Chen, Xie, Li, Doll{\'a}r, and
  Girshick}]{he2022mae}
He, K.; Chen, X.; Xie, S.; Li, Y.; Doll{\'a}r, P.; and Girshick, R.
  2022{\natexlab{b}}.
\newblock Masked autoencoders are scalable vision learners.
\newblock In \emph{CVPR}.

\bibitem[{He et~al.(2017)He, Gkioxari, Dollar, and Girshick}]{he2017maskrcnn}
He, K.; Gkioxari, G.; Dollar, P.; and Girshick, R. 2017.
\newblock Mask R-CNN.
\newblock In \emph{ICCV}.

\bibitem[{He et~al.(2016)He, Zhang, Ren, and Sun}]{he2016deep}
He, K.; Zhang, X.; Ren, S.; and Sun, J. 2016.
\newblock Deep residual learning for image recognition.
\newblock In \emph{CVPR}.

\bibitem[{Hoefler et~al.(2021)Hoefler, Alistarh, Ben-Nun, Dryden, and
  Peste}]{hoefler2021sparsity}
Hoefler, T.; Alistarh, D.; Ben-Nun, T.; Dryden, N.; and Peste, A. 2021.
\newblock Sparsity in deep learning: Pruning and growth for efficient inference
  and training in neural networks.
\newblock \emph{The Journal of Machine Learning Research}.

\bibitem[{Houlsby et~al.(2019)Houlsby, Giurgiu, Jastrzebski, Morrone,
  De~Laroussilhe, Gesmundo, Attariyan, and Gelly}]{houlsby2019parameter}
Houlsby, N.; Giurgiu, A.; Jastrzebski, S.; Morrone, B.; De~Laroussilhe, Q.;
  Gesmundo, A.; Attariyan, M.; and Gelly, S. 2019.
\newblock Parameter-efficient transfer learning for NLP.
\newblock In \emph{ICML}.

\bibitem[{Hu et~al.(2022)Hu, Wallis, Allen-Zhu, Li, Wang, Wang, Chen
  et~al.}]{hu2022lora}
Hu, E.~J.; Wallis, P.; Allen-Zhu, Z.; Li, Y.; Wang, S.; Wang, L.; Chen, W.;
  et~al. 2022.
\newblock LoRA: Low-Rank Adaptation of Large Language Models.
\newblock In \emph{ICLR}.

\bibitem[{Jia et~al.(2022)Jia, Tang, Chen, Cardie, Belongie, Hariharan, and
  Lim}]{jia2022vpt}
Jia, M.; Tang, L.; Chen, B.-C.; Cardie, C.; Belongie, S.; Hariharan, B.; and
  Lim, S.-N. 2022.
\newblock Visual Prompt Tuning.
\newblock In \emph{ECCV}.

\bibitem[{Kendall, Gal, and Cipolla(2018)}]{kendall2018multi}
Kendall, A.; Gal, Y.; and Cipolla, R. 2018.
\newblock Multi-task learning using uncertainty to weigh losses for scene
  geometry and semantics.
\newblock In \emph{CVPR}.

\bibitem[{Lan et~al.(2020)Lan, Chen, Goodman, Gimpel, Sharma, and
  Soricut}]{lan2020albert}
Lan, Z.; Chen, M.; Goodman, S.; Gimpel, K.; Sharma, P.; and Soricut, R. 2020.
\newblock ALBERT: A Lite BERT for Self-supervised Learning of Language
  Representations.
\newblock In \emph{ICLR}.

\bibitem[{Lee et~al.(2022)Lee, Chang, Jiang, Zhang, Tu, and
  Liu}]{lee2021vitgan}
Lee, K.; Chang, H.; Jiang, L.; Zhang, H.; Tu, Z.; and Liu, C. 2022.
\newblock Vitgan: Training gans with vision transformers.
\newblock In \emph{ICLR}.

\bibitem[{Lian et~al.(2022)Lian, Zhou, Feng, and Wang}]{lian2022ssf}
Lian, D.; Zhou, D.; Feng, J.; and Wang, X. 2022.
\newblock Scaling \& Shifting Your Features: A New Baseline for Efficient Model
  Tuning.
\newblock In \emph{NeurIPS}.

\bibitem[{Lin et~al.(2020)Lin, Stich, Barba, Dmitriev, and
  Jaggi}]{lin2020dynamic}
Lin, T.; Stich, S.~U.; Barba, L.; Dmitriev, D.; and Jaggi, M. 2020.
\newblock Dynamic model pruning with feedback.
\newblock In \emph{ICLR}.

\bibitem[{Lin et~al.(2017)Lin, Dollar, Girshick, He, Hariharan, and
  Belongie}]{lin2017fpn}
Lin, T.-Y.; Dollar, P.; Girshick, R.; He, K.; Hariharan, B.; and Belongie, S.
  2017.
\newblock Feature Pyramid Networks for Object Detection.
\newblock In \emph{CVPR}.

\bibitem[{Lin et~al.(2014)Lin, Maire, Belongie, Hays, Perona, Ramanan,
  Doll{\'a}r, and Zitnick}]{lin2014coco}
Lin, T.-Y.; Maire, M.; Belongie, S.; Hays, J.; Perona, P.; Ramanan, D.;
  Doll{\'a}r, P.; and Zitnick, C.~L. 2014.
\newblock Microsoft coco: Common objects in context.
\newblock In \emph{ECCV}.

\bibitem[{Lin et~al.(2022)Lin, Liu, Zhang, Hu, Zheng, Lin, and
  Cao}]{lin2022could}
Lin, Y.; Liu, Z.; Zhang, Z.; Hu, H.; Zheng, N.; Lin, S.; and Cao, Y. 2022.
\newblock Could Giant Pretrained Image Models Extract Universal
  Representations?
\newblock In \emph{NeurIPS}.

\bibitem[{Liu, Simonyan, and Yang(2019)}]{liu2018darts}
Liu, H.; Simonyan, K.; and Yang, Y. 2019.
\newblock Darts: Differentiable architecture search.
\newblock In \emph{ICLR}.

\bibitem[{Liu et~al.(2019)Liu, Ott, Goyal, Du, Joshi, Chen, Levy, Lewis,
  Zettlemoyer, and Stoyanov}]{liu2019roberta}
Liu, Y.; Ott, M.; Goyal, N.; Du, J.; Joshi, M.; Chen, D.; Levy, O.; Lewis, M.;
  Zettlemoyer, L.; and Stoyanov, V. 2019.
\newblock Roberta: A robustly optimized bert pretraining approach.
\newblock \emph{arXiv preprint arXiv:1907.11692}.

\bibitem[{Liu et~al.(2022)Liu, Hu, Lin, Yao, Xie, Wei, Ning, Cao, Zhang, Dong
  et~al.}]{liu2022swin}
Liu, Z.; Hu, H.; Lin, Y.; Yao, Z.; Xie, Z.; Wei, Y.; Ning, J.; Cao, Y.; Zhang,
  Z.; Dong, L.; et~al. 2022.
\newblock Swin transformer v2: Scaling up capacity and resolution.
\newblock In \emph{CVPR}.

\bibitem[{Liu et~al.(2021)Liu, Lin, Cao, Hu, Wei, Zhang, Lin, and
  Guo}]{liu2021swinv1}
Liu, Z.; Lin, Y.; Cao, Y.; Hu, H.; Wei, Y.; Zhang, Z.; Lin, S.; and Guo, B.
  2021.
\newblock Swin Transformer: Hierarchical Vision Transformer using Shifted
  Windows.
\newblock In \emph{ICCV}.

\bibitem[{Nathan~Silberman and Fergus(2012)}]{silber2012nyuv2}
Nathan~Silberman, P.~K., Derek~Hoiem; and Fergus, R. 2012.
\newblock Indoor Segmentation and Support Inference from RGBD Images.
\newblock In \emph{ECCV}.

\bibitem[{Qiao et~al.(2022)Qiao, Ou, Zhang, Chen, Yao, Deng, Tan, Huang, and
  Chen}]{qiao2022survey}
Qiao, S.; Ou, Y.; Zhang, N.; Chen, X.; Yao, Y.; Deng, S.; Tan, C.; Huang, F.;
  and Chen, H. 2022.
\newblock Reasoning with Language Model Prompting: A Survey.
\newblock \emph{arXiv preprint arXiv:2212.09597}.

\bibitem[{Ranftl, Bochkovskiy, and Koltun(2021)}]{ranftl2021vision}
Ranftl, R.; Bochkovskiy, A.; and Koltun, V. 2021.
\newblock Vision transformers for dense prediction.
\newblock In \emph{ICCV}.

\bibitem[{Ridnik et~al.(2021)Ridnik, Ben-Baruch, Noy, and
  Zelnik-Manor}]{ridnik2021imagenet}
Ridnik, T.; Ben-Baruch, E.; Noy, A.; and Zelnik-Manor, L. 2021.
\newblock Imagenet-21k pretraining for the masses.
\newblock \emph{arXiv preprint arXiv:2104.10972}.

\bibitem[{Stickland and Murray(2019)}]{stickland2019pal}
Stickland, A.~C.; and Murray, I. 2019.
\newblock Bert and pals: Projected attention layers for efficient adaptation in
  multi-task learning.
\newblock In \emph{ICML}.

\bibitem[{Sun et~al.(2017)Sun, Shrivastava, Singh, and
  Gupta}]{sun2017revisiting}
Sun, C.; Shrivastava, A.; Singh, S.; and Gupta, A. 2017.
\newblock Revisiting unreasonable effectiveness of data in deep learning era.
\newblock In \emph{ICCV}.

\bibitem[{Tam et~al.(2021)Tam, Menon, Bansal, Srivastava, and
  Raffel}]{tam2021adapet}
Tam, D.; Menon, R.~R.; Bansal, M.; Srivastava, S.; and Raffel, C. 2021.
\newblock Improving and Simplifying Pattern Exploiting Training.
\newblock In \emph{EMNLP}.

\bibitem[{Tan, Pang, and Le(2020)}]{tan2020bifpn}
Tan, M.; Pang, R.; and Le, Q.~V. 2020.
\newblock Efficientdet: Scalable and efficient object detection.
\newblock In \emph{CVPR}.

\bibitem[{Touvron et~al.(2021)Touvron, Cord, Douze, Massa, Sablayrolles, and
  J{\'e}gou}]{touvron2021deit}
Touvron, H.; Cord, M.; Douze, M.; Massa, F.; Sablayrolles, A.; and J{\'e}gou,
  H. 2021.
\newblock Training data-efficient image transformers \& distillation through
  attention.
\newblock In \emph{ICML}.

\bibitem[{Vandenhende et~al.(2021)Vandenhende, Georgoulis, Van~Gansbeke,
  Proesmans, Dai, and Van~Gool}]{vandenhende2021multi}
Vandenhende, S.; Georgoulis, S.; Van~Gansbeke, W.; Proesmans, M.; Dai, D.; and
  Van~Gool, L. 2021.
\newblock Multi-task learning for dense prediction tasks: A survey.
\newblock \emph{IEEE transactions on pattern analysis and machine
  intelligence}.

\bibitem[{Wang et~al.(2022)Wang, Yao, Chen, Cai, He, and
  Liu}]{wang2021crossformer}
Wang, W.; Yao, L.; Chen, L.; Cai, D.; He, X.; and Liu, W. 2022.
\newblock Crossformer: A versatile vision transformer based on cross-scale
  attention.
\newblock In \emph{ICLR}.

\bibitem[{Xie et~al.(2021)Xie, Wang, Yu, Anandkumar, Alvarez, and
  Luo}]{xie2021segformer}
Xie, E.; Wang, W.; Yu, Z.; Anandkumar, A.; Alvarez, J.~M.; and Luo, P. 2021.
\newblock SegFormer: Simple and Efficient Design for Semantic Segmentation with
  Transformers.
\newblock In \emph{NeurIPS}.

\bibitem[{Xie et~al.(2019)Xie, Zheng, Liu, and Lin}]{xie2018snas}
Xie, S.; Zheng, H.; Liu, C.; and Lin, L. 2019.
\newblock SNAS: stochastic neural architecture search.
\newblock In \emph{ICLR}.

\bibitem[{Xie et~al.(2022{\natexlab{a}})Xie, Geng, Hu, Zhang, Hu, and
  Cao}]{xie2022revealing}
Xie, Z.; Geng, Z.; Hu, J.; Zhang, Z.; Hu, H.; and Cao, Y. 2022{\natexlab{a}}.
\newblock Revealing the dark secrets of masked image modeling.
\newblock \emph{arXiv preprint arXiv:2205.13543}.

\bibitem[{Xie et~al.(2022{\natexlab{b}})Xie, Geng, Hu, Zhang, Hu, and
  Cao}]{xie2023mmdepth}
Xie, Z.; Geng, Z.; Hu, J.; Zhang, Z.; Hu, H.; and Cao, Y. 2022{\natexlab{b}}.
\newblock Revealing the Dark Secrets of Masked Image Modeling.
\newblock \emph{arXiv preprint arXiv:2205.13543}.

\bibitem[{Yin et~al.(2020)Yin, Kim, Oh, Wang, Serrano, Seo, and
  Choi}]{yin2020sooner}
Yin, S.; Kim, K.-H.; Oh, J.; Wang, N.; Serrano, M.; Seo, J.-S.; and Choi, J.
  2020.
\newblock The sooner the better: Investigating structure of early winning
  lottery tickets.

\bibitem[{Yu and Huang(2019)}]{yu2019autoslim}
Yu, J.; and Huang, T. 2019.
\newblock Autoslim: Towards one-shot architecture search for channel numbers.
\newblock \emph{arXiv preprint arXiv:1903.11728}.

\bibitem[{Zhai et~al.(2022)Zhai, Kolesnikov, Houlsby, and
  Beyer}]{zhai2022scaling}
Zhai, X.; Kolesnikov, A.; Houlsby, N.; and Beyer, L. 2022.
\newblock Scaling vision transformers.
\newblock In \emph{CVPR}.

\bibitem[{Zhai et~al.(2019)Zhai, Puigcerver, Kolesnikov, Ruyssen, Riquelme,
  Lucic, Djolonga, Pinto, Neumann, Dosovitskiy et~al.}]{zhai2019large}
Zhai, X.; Puigcerver, J.; Kolesnikov, A.; Ruyssen, P.; Riquelme, C.; Lucic, M.;
  Djolonga, J.; Pinto, A.~S.; Neumann, M.; Dosovitskiy, A.; et~al. 2019.
\newblock A large-scale study of representation learning with the visual task
  adaptation benchmark.
\newblock \emph{arXiv preprint arXiv:1910.04867}.

\bibitem[{Zhang et~al.(2022)Zhang, Li, Chen, Deng, Bi, Tan, Huang, and
  Chen}]{zhang2022dart}
Zhang, N.; Li, L.; Chen, X.; Deng, S.; Bi, Z.; Tan, C.; Huang, F.; and Chen, H.
  2022.
\newblock Differentiable Prompt Makes Pre-trained Language Models Better
  Few-shot Learners.
\newblock In \emph{ICLR}.

\bibitem[{Zhang, Zhou, and Liu(2022)}]{zhang2022noah}
Zhang, Y.; Zhou, K.; and Liu, Z. 2022.
\newblock Neural prompt search.
\newblock \emph{arXiv preprint arXiv:2206.04673}.

\bibitem[{Zhou et~al.(2017)Zhou, Zhao, Puig, Fidler, Barriuso, and
  Torralba}]{zhou2017ade20k}
Zhou, B.; Zhao, H.; Puig, X.; Fidler, S.; Barriuso, A.; and Torralba, A. 2017.
\newblock Scene parsing through ade20k dataset.
\newblock In \emph{CVPR}.

\bibitem[{Zhu and Gupta(2017)}]{zhu2017prune}
Zhu, M.; and Gupta, S. 2017.
\newblock To prune, or not to prune: exploring the efficacy of pruning for
  model compression.
\newblock \emph{arXiv preprint arXiv:1710.01878}.

\bibitem[{Zoph and Le(2017)}]{zoph2016neural}
Zoph, B.; and Le, Q.~V. 2017.
\newblock Neural architecture search with reinforcement learning.
\newblock In \emph{ICLR}.

\end{thebibliography}
}

\end{document}


\title{Selective Feature Adapter for Dense Vision Transformers}

\author{First Author\\
Institution1\\
Institution1 address\\
{\tt\small firstauthor@i1.org}
\and
Second Author\\
Institution2\\
First line of institution2 address\\
{\tt\small secondauthor@i2.org}
}

\maketitle
\ificcvfinal\thispagestyle{empty}\fi

\section{More Experimental Results}

\subsection{Comparison study with Vit-adapter}

Table~\ref{table:vit} shows that our method outperforms vit-adapter~\cite{chen2022vitadapter} on all backbones with fewer trainable parameters.
With more trainable backbone parameters, our method significantly outperforms vit-adapter.

\subsection{Selection criteria for internal adapter $\mathcal{F}_i$.} 
While our method selects parameters with the highest gradient as stated in Eqn.(4) and Eqn.(4) in the paper, there are other alternatives for selecting parameters to train. 
Firstly, we compare multiple selection metrics including Random and Magnitude proposed in~\cite{he-etal-2022-sparseadapter} to $\mathcal{F}_i$. 
Random selection assigns a random score z $\sim$ Uniform(0, 1) to each parameter and chooses the top $\beta_i$. While the Magnitude selection chooses those weights with the large exact magnitude as the trainable weights. Surprisingly, Table~\ref{table:metrics} shows random selection is also able to perform pretty well, while using magnitude gives a better performance, from 53.7 to 54.0 


In Tab.~\ref{tab:metrics}, 
the "Layer-wise" selection involves choosing parameters with the highest scores based on Eqn.(4) from the paper, allocating an equal budget to each layer. Let our budget portion be $\beta_i$, and let there be $l$ layers. Each layer receives a $\beta_i/l$ portion for selecting trainable parameters. This method yields lower performance (54.3) compared to our unconstrained strategy (54.7), underscoring the efficacy of the SFA's automatic selection over manual constraints.

\begin{table}[!h] 
  \caption{Comparison study on ADE20K with multi-scale mIoU. The trainable parameters ratio is shown in parentheses.} 
    \centering
         \resizebox{\linewidth}{!}{%
      \begin{tabular}{l|ccc|cc}
        \toprule
        Method & Param \% & S-B-1K & S-B-22K & Param \% & S-L-22K \\
        \midrule
        Frozen Vit~\cite{lin2022could} & 0 & 43.1 & 44.7 & 0 & 46.2 \\
        Vit-adapter~\cite{chen2022vitadapter} & 15.9 & 45.8 & 47.4 & 12.0 & 47.8 \\
        SFA-L (ours) & 4.8 & \textbf{46.9} & \textbf{47.8} & 4.5 & \textbf{48.3}  \\
        \midrule
        Fully fine-tune  & 100 & 48.2 & 49.4 & 100 & 50.1 \\
        SFA-B (ours) & 19.1 & 47.6 & 48.6 & 18.6 & 49.4 \\
        \bottomrule
      \end{tabular}}
\vspace{-5pt}
\label{table:vit}
\end{table}

\begin{table}[h] 
  \caption{Comparisons of different internal adapter selection criteria. mIoU scores on ADE20K are reported.} 
  \vspace{-5pt}
    \centering
     \resizebox{\linewidth}{!}{%
      \begin{tabular}{c|cc|cc}
        \toprule
        & \multicolumn{2}{c|}{Selection Criteria} & \multicolumn{2}{c}{Selection Manners}  \\
        Frozen~\cite{lin2022could} & Random & Magnitude & Layer-wise   & $\mathcal{F}_i$ (ours) \\
        \midrule
        53.2 & 53.7 & 54.0 & 54.3 & 54.7 \\
        \bottomrule
      \end{tabular}
      }
\label{table:metrics}
\end{table}

{\small
\bibliographystyle{ieee_fullname}
\bibliography{egbib}
}